\documentclass{article}

\usepackage{arxiv}
\usepackage{natbib}
\usepackage{makecell}
\usepackage{multibib}
\usepackage{graphicx}
\usepackage{tabularx}

\usepackage{times}
\usepackage{latexsym}
\usepackage[T1]{fontenc}
\usepackage{subfloat}
\usepackage{subcaption}
\usepackage[utf8]{inputenc}

\usepackage{microtype}

\usepackage{inconsolata}

\usepackage[utf8]{inputenc} 
\usepackage[T1]{fontenc}    
\usepackage{hyperref}       
\usepackage{url}            
\usepackage{booktabs}       
\usepackage{amsfonts}       
\usepackage{nicefrac}       
\usepackage{microtype}      
\usepackage{lipsum}		
\usepackage{graphicx}
\usepackage{natbib}
\usepackage{doi}
\usepackage{xcolor}
\usepackage{natbib}
\usepackage{array}
\usepackage{makecell}

\usepackage[utf8]{inputenc} 
\usepackage[T1]{fontenc}    
\usepackage{hyperref}       
\usepackage{url}            
\usepackage{booktabs}       
\usepackage{amsfonts}       
\usepackage{nicefrac}       
\usepackage{microtype}      
\usepackage{xcolor}         

\usepackage{graphicx}

\title{CARBD-Ko: A Contextually Annotated Review Benchmark Dataset for Aspect-Level Sentiment Classification in Korean}

\author{Dongjun Jang \\
  Department of Linguistics \\
  Seoul National University \\
  \texttt{qwer4107@snu.ac.kr} \\
  \And
  Jean Seo \\
  Department of Linguistics \\
  Seoul National University \\
  \texttt{seemdog@snu.ac.kr} \\
  \And
  Sungjoo Byun \\
  Department of Linguistics \\
  Seoul National University \\
  \texttt{byunsj@snu.ac.kr} \\
  \And
  Taekyoung Kim \\
  Graduate School of Data Science \\
  Seoul National University \\
  \texttt{taekyoung@snu.ac.kr} \\
  \And
  Minseok Kim \\
  Department of Linguistics \\
  Seoul National University \\
  \texttt{snumin44@snu.ac.kr} \\
  \And
  Hyopil Shin \\
  Department of Linguistics \\
  Seoul National University \\
  \texttt{hpshin@snu.ac.kr} \\
}
\renewcommand{\shorttitle}{\textit{arXiv} Template}

\hypersetup{
pdftitle={A template for the arxiv style},
pdfsubject={q-bio.NC, q-bio.QM},
pdfauthor={David S.~Hippocampus, Elias D.~Striatum},
pdfkeywords={First keyword, Second keyword, More},
}

\begin{document}
\maketitle

\begin{abstract}
This paper explores the challenges posed by aspect-based sentiment classification (ABSC) within pretrained language models (PLMs), with a particular focus on contextualization and hallucination issues. In order to tackle these challenges, we introduce CARBD-Ko (a Contextually Annotated Review Benchmark Dataset for Aspect-Based Sentiment Classification in Korean), a benchmark dataset that incorporates aspects and dual-tagged polarities to distinguish between aspect-specific and aspect-agnostic sentiment classification. The dataset consists of sentences annotated with specific aspects, aspect polarity, aspect-agnostic polarity, and the intensity of aspects. To address the issue of dual-tagged aspect polarities, we propose a novel approach employing a Siamese Network. Our experimental findings highlight the inherent difficulties in accurately predicting dual-polarities and underscore the significance of contextualized sentiment analysis models. The CARBD-Ko dataset serves as a valuable resource for future research endeavors in aspect-level sentiment classification.
\end{abstract}


\section{Introduction}
The effectiveness of various pretrained language models, including BERT \citep{devlin2018bert}, XLNet \citep{yang2019xlnet}, BART \citep{lewis-etal-2020-bart}, and GPT-3, in sentiment classification, a significant downstream task, has been extensively studied. Current research in sentiment classification often focuses on identifying sentiment polarities at the aspect level, leading to the emergence of aspect-based sentiment classification (ABSC). Many studies have achieved impressive results and introduced innovative approaches to tackle the ABSC task. For instance, \citet{sun2019utilizing} utilized BERT to transform ABSC tasks into sentence-pair classification, which has influenced subsequent methodologies \citep{hu-etal-2022-unimse}. Additionally, generative models like BART \citep{lewis-etal-2020-bart} have been employed by \citet{yan-etal-2021-unified} to convert ABSC tasks into sequence-to-sequence tasks, enabling the prediction of token sequences representing identified aspects and associated sentiments. Furthermore, \citet{li-etal-2021-learning-implicit} reframed ABSC tasks as masked language modeling tasks, effectively bridging the performance gap between pre-training and ABSC tasks.
\begin{figure*}[ht]
  \includegraphics[width=\textwidth,height=2.5cm]{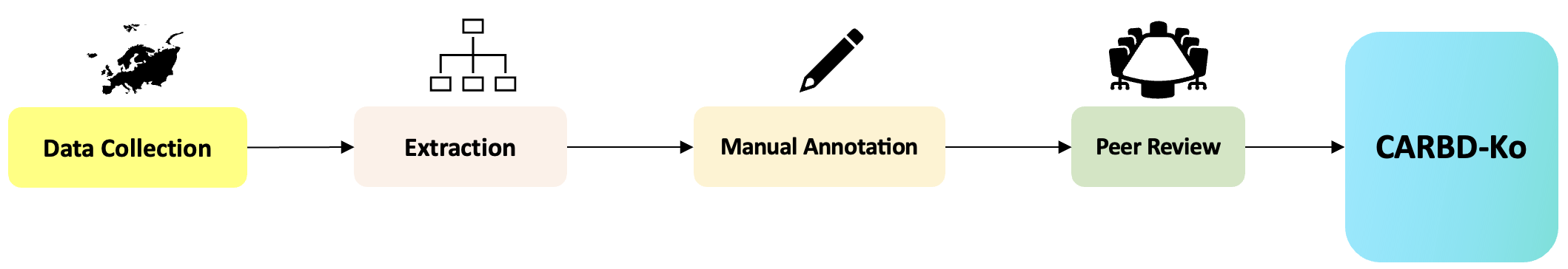}
  \caption{The figure provides an overview of the pipeline used to construct CARBD-Ko. It involves four steps, starting with the collection of comments from diverse domains. Next, aspect-opinion pairs are extracted from the comments. The pipeline also includes the manual annotation of both aspect-agnostic and aspect polarity, and intensity. To ensure objectivity, a peer review stage is incorporated. Overall, this pipeline enables a comprehensive sentiment analysis of the comment data in CARBD-Ko}
  \label{pipeline}
\end{figure*}

Despite numerous attempts to address aspect-level sentiment classification, the primary focus has been on improving aspect-level sentiment polarity performance through specialized datasets and training methodologies. However, it is equally crucial for models to predict not only the in-context polarity of aspects but also their aspect polarity. Unfortunately, both polarities are frequently overlooked when devising fine-tuning strategies. Existing sentiment classification research has largely ignored the contextualization issue inherent to aspects. For instance, if we restrict the sentiment label to only negative for the statement "Room service was good, but that's all!" or categorize it as positive solely for the "room service" segment, it limits the model's capacity to address the problem of hallucination in accurately predicting aspect sentiment across different contexts. Consequently, there is a need to annotate sentiment values that are both context-autonomous and context-dependent, in a contextual dimension. 

In this study, we take these factors into consideration and introduce CARBD-Ko (A Korean Contextually Annotated Review Benchmark Dataset for Aspect-Based Sentiment Classification), a unique benchmark dataset characterized by both aspect and aspect-agnostic polarities. Each sentence in CARBD-Ko is annotated with a specific aspect and its accompanying sentiment details. The double-tagged aspect polarity of CARBD-Ko distinguishes it from existing datasets. Additionally, we propose a novel modeling approach that employs a Siamese Network to handle this double-tagged polarity. Furthermore, we highlight the performance of our model on CARBD-Ko and provide a comparative analysis with pre-existing sentiment classification benchmark datasets.

Our study makes a valuable contribution by detailing the construction process of the CARBD-Ko dataset, which is characterized by dual-polarity annotations. Moreover, it presents a fresh perspective through a novel modeling approach using Siamese Networks following \citet{bromley1993signature}. Additionally, it provides insights into the potential challenges and future directions of aspect-agnostic sentiment analysis, particularly within the scope of aspect-level sentiment classification.
\section{Related Work}
\textbf{Aspect-based Sentiment Analysis}

Aspect-Based Sentiment Classification (ABSC) allows a detailed sentiment analysis by determining the sentiment polarity at the aspect-level. Oftentimes, the overall text sentiment doesn't align with sentiments pertaining to all its constituent aspects. As such, it's vital to extract these aspects and assign sentiment polarity individually for comprehensive sentiment analysis \citep{liu2012sentiment}. Recent works aim to improve aspect extraction and boost ABSC performance. For instance, \citet{liang2021beta} employ aspect-sensitive terms and their weights to establish an aspect-aware graph convolutional structure. Similarly, \citet{li2021learning} introduce large-scale domain-specific annotated corpora and Supervised Contrastive Pretraining for ABSA to capture implicit sentiment nuances.

\textbf{Benchmark for Low-Resource Languages}
Benchmarks play a crucial role in advancing research in Natural Language Processing (NLP), highlighting the need for benchmarks in low-resource languages. Several benchmarks have been developed for different languages to drive progress in NLP tasks. For instance, IndoNLG is an Indonesian benchmark specifically designed for natural language generation tasks, covering six NLG tasks \citep{cahyawijaya2021indonlg}. Additionally, the Flores-101 evaluation benchmark includes data from 101 low-resource languages, promoting research and development in these languages \citep{goyal2022flores}. In line with these efforts, we introduce a Korean benchmark for aspect-based sentiment analysis, addressing the low-resource nature of the Korean language and providing a standardized evaluation framework to advance sentiment analysis research in low-resource languages.

\textbf{The Siamese Network}
The Siamese Network was first introduce by \citep{bromley1993signature}, which Network consist of two identical sub-networks joined at their outputs. The Siamese Network, inspired by the success of Sentence-BERT \citep{reimers2019sentencebert}, has gained popularity in the field of Natural Language Processing, especially in sentiment analysis. Researchers have leveraged the Siamese Network in various ways to enhance sentiment analysis models. For example, \citet{choudhary2018emotions, choudhary2018sentiment} utilized the Siamese Network to improve representations for resource-poor languages. \citet{huang2018siamese} developed a supervised topic modeling model using additional sentiment labels. Additionally, \citet{zhang2022kasn} implemented the Siamese Network to train models using external sentiment knowledge. These studies demonstrate the versatility and effectiveness of the Siamese Network in sentiment analysis tasks. 

\begin{figure*}
  \includegraphics[width=\textwidth,height=8cm]{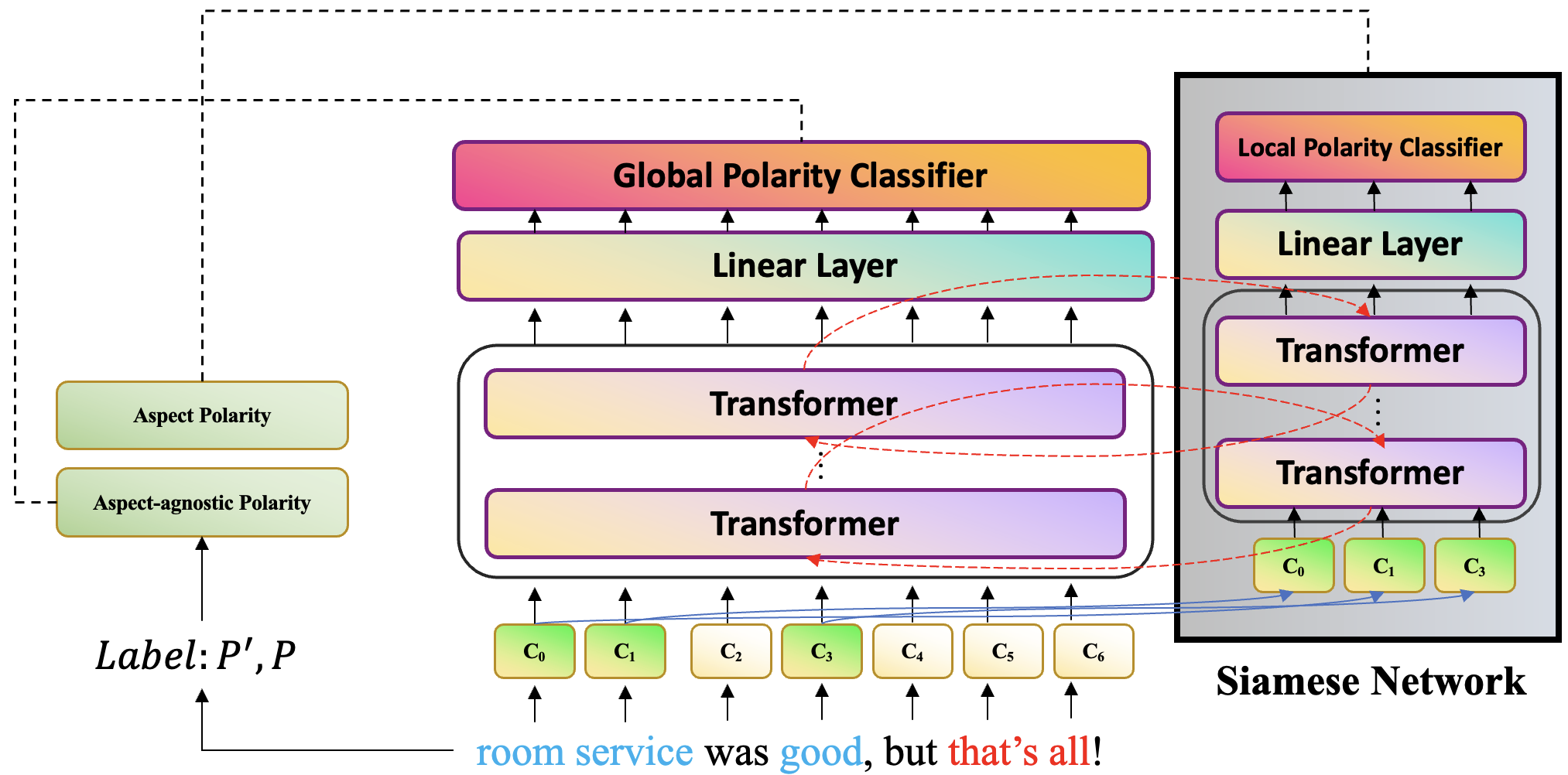}
  \caption{The Operation of \textbf{Siamese Network} for training CARBD-Ko. While fine-tuning Transformer-based model on Sentiment Classification, the aspect-opinion token pair is passed through the Siamese Network to reduce the bias of the polarity value due to contextualization. The model is trained by simultaneously learning aspect-agnostic polarity and aspect polarity.
}
\label{modeling}
\end{figure*}

\section{CARBD-Ko}
We present CARBD-Ko (A Contextually Annotated Review Benchmark Dataset for Aspect-Based Sentiment Classification in Korean), a challenge dataset constructed from various review comments. The dataset is created through a series of four main steps, resulting in 10,586 instances for training and nearly 3,000 sets for validation (Figure \ref{pipeline}). Our goal is to continually expand the CARBD-Ko dataset to facilitate future research investigations.

\subsection{Data Collection}
We collected publicly available comments from anonymous users across diverse domains, including movies and shopping.

\subsection{Aspect and Opinion Extraction}
We developed an internal algorithm that extracts aspects and opinions from the sentences. This extraction process is guided by the definition of Dependency Relationship and Part-Of-Speech provided by Stanza\footnote{\url{https://stanfordnlp.github.io/stanza/}}. In this process, an aspect $x_\alpha$ represents a specific target or topic within a sentence $x$, while an opinion $x_\omega$ is a phrase indicating feelings towards the aspect $x_\alpha$.

\[
x_\alpha, x_\omega = stanza(x)
\]

\subsection{Manual Annotation}
A team of five annotators manually assigns Aspect-Agnostic Polarity $P(x_\alpha, x_\omega)$, Aspect Polarity $P'(x_\alpha, x_\omega)$, and the Intensity of Aspect Polarity $I'$. The values of $P$ and $P'$ fall within the range of {-1, 0, 1}. This can be represented as follows:
\[
{P(x_\alpha, x_\omega), P'(x_\alpha, x_\omega)} \in \{-1, 0, 1\},
\]
where $-1$ denotes a negative sentiment, $0$ represents a neutral sentiment, and $1$ signifies a positive sentiment. Lastly, $I'$ represents the intensity of the sentiment, which is evaluated using a 7-point Likert scale:
\[
I'(x_\alpha, x_\omega) \in \{-3, -2, -1, 0, 1, 2, 3\}.
\]
Here, $I' = -3$ indicates a highly intense negative sentiment, $I' = -2$ represents a moderately intense negative sentiment, $I' = -1$ signifies a mildly intense negative sentiment, $I' = 0$ denotes a neutral intensity, $I' = 1$ reflects a mildly intense positive sentiment, $I' = 2$ shows a moderately intense positive sentiment, and $I' = 3$ demonstrates a highly intense positive sentiment. These intensity levels provide a fine-grained understanding of the strength of sentiment associated with the aspects analyzed in our study.

\subsection{Peer Review Stage} 
To enhance the objectivity and precision of sentiment value assessment, the dataset is subjected to a peer review stage. Four annotators are organized into two groups, and each group independently annotates a randomly shuffled subset of the dataset. This process helps ensure consistency and minimize biases in sentiment value assignments.

\subsection{Dataset Overview and Statistical Analysis}

\begin{table}[ht]

\centering
\small\begin{tabularx}{0.5\linewidth}{X*{5}{c}}
\hline
    \toprule
        \textbf{Polarity} &\textbf{$-1$}  & \textbf{$0$}  & \textbf{$1$} \\
    \hline
    \hline
    \midrule
        \textbf{$P(x_\alpha, x_\omega)$} & 5121 &  516 & 4949 \\
        \textbf{$P'(x_\alpha, x_\omega)$} & 4986 & 1203 & 4397 \\
        \hline
        \textbf{Total} & 10107 & 1719 & 9346\\

    \midrule
\hline
\end{tabularx}
\caption{Label Distribution Per Polarity Type of Train Dataset of CARBD-Ko}
\label{overview}
\end{table}

\begin{table}[ht]

\centering
\small\begin{tabularx}{0.5\linewidth}{X*{5}{c}}
\hline
    \toprule
        \textbf{Polarity} &\textbf{$-1$}  & \textbf{$0$}  & \textbf{$1$} \\
    \hline
    \hline
    \midrule
        \textbf{$P(x_\alpha, x_\omega)$} & 1597 &  3 & 1400 \\
        \textbf{$P'(x_\alpha, x_\omega)$} & 1541 & 124 & 1335 \\
        \hline
        \textbf{Total} & 3138 & 127 & 2735\\

    \midrule
\hline
\end{tabularx}
\caption{Label Distribution Per Polarity Type of Validation Dataset of CARBD-Ko}
\label{overview2}
\end{table}

Tables \ref{overview} and \ref{overview2} offer a succinct depiction of label distribution within the training and validation datasets, respectively. The labels -1, 0, and 1 signify negative, neutral, and positive polarities, respectively, and the numerical values within each cell denote the dataset sizes corresponding to these polarity categories. Additionally, Table \ref{overview3} provides essential statistical metrics for both the training and validation datasets, including measures such as mean, minimum, maximum, and standard deviation. These statistical insights provide a comprehensive overview of the CARBD-Ko dataset's characteristics, serving as a valuable resource for understanding its composition.

\begin{table}

\centering
\small\begin{tabularx}{0.5\linewidth}{X*{5}{c}}
\hline
    \toprule
        \textbf{Dataset} &\textbf{Mean}  & \textbf{Min}  & \textbf{Max} & \textbf{Std} \\
    \hline
    \hline
    \midrule
        \textbf{Training} & 31.26 &  5 & 177 & 19.63 \\
        \textbf{Validation} & 29.72 & 5 & 177 & 18.57 \\

    \midrule
\hline
\end{tabularx}
\caption{Descriptive Statistics of Training and Validation Datasets}
\label{overview3}
\end{table}

\section{Simultaneous Learning of Dual-Polarities via Siamese Network}
To effectively capture and learn the dual-polarities, we propose a collective learning approach using a Siamese Network (Figure \ref{modeling}), which architecture is similar to original Siamese Network \citep{bromley1993signature}. This network incorporates shared parameters that are updated during each back-propagation cycle, enabling the model to effectively integrate both aspects of polarity. The Siamese Network undergoes training using a combined loss function that integrates the losses from two classifiers, namely the Global Polarity Classifier and the Local Polarity Classifier. We hypothesize that this training procedure enables the model to effectively capture and integrate sentiment information in the given context.

\[
Joint(L) = L(P', \hat{P'}) + L(P, \hat{P})
\]

\section{Experiment}
Our experimental process focuses on evaluating the performance of four prominent Korean Encoder-based Pretrained language models (ko-electra \citep{park2020koelectra}, kr-electra \citep{kr-electra}, kc-electra \citep{lee2021kcelectra}, kr-bert \citep{lee2020krbert} along with the XLM-roberta-base model \citep{conneau2019unsupervised}. Initially, we assess their performance on the NSMC Benchmark\footnote{\url{https://github.com/e9t/nsmc}}, a Korean sentiment analysis task, to gain insights into the challenges of sentiment analysis. Subsequently, we evaluate these models on the CARBD-Ko dataset using a Siamese Network.

\begin{table}

\centering
\small\begin{tabularx}{\linewidth}{X*{5}{c}}
\hline
    \toprule
        \textbf{Model} & \textbf{NSMC} &\textbf{$P$}  & \textbf{$P'$}  & \textbf{$P \& P'$} \\
    \hline
    \hline
    \midrule
        \textbf{ko-electra} & 90.63 &  70.9& 76.0 &65.2\\
        \textbf{kr-electra} & 91.17 & \textbf{79.8}& \textbf{85.5} & \textbf{74.1}\\
        \textbf{kc-electra} & 91.97 & \textbf{79.8}& 83.9 & 73.2\\
        \textbf{xlm-roberta-base} & 89.49 & 79.3& 85.8 & 73.9\\
        \textbf{kr-bert}& 90.1 &  70.9 & 75.4 & 64.9\\
    \midrule
\hline
\end{tabularx}
\caption{Accuracy of Performance Evaluation of Models on NSMC and CARBD-Ko Benchmarks}
\label{performance}
\end{table}

\subsection{Settings}
To achieve optimal performance, we conduct hyper-parameter optimization by adjusting the learning rate within the range of 1e-5 to 5e-5 and increasing the number of training epochs from 3 to 10.

\section{Results}
The experimental results presented in Table \ref{performance} indicate that all models perform well on the NSMC benchmark. However, when evaluating the models on the CARBD-Ko dataset with the Siamese Network, we observe certain challenges in aspect-agnostic sentiment prediction. The accuracy for predicting the Aspect Polarity ($P$) is noticeably lower compared to predicting the Aspect-Agnostic Polarity ($P'$). Moreover, the accuracy further decreases when considering the simultaneous prediction of both polarities ($P\&P'$). This suggests that language models still struggle to dynamically adjust sentiment values based on the context during sentiment analysis. These findings support the necessity of expanding the CARBD-Ko benchmark and emphasize the importance of context in aspect-based sentiment classification, calling for further research in this area.

\section{Conclusion}

This work proposes a novel approach towards addressing context-dependency in aspect-level sentiment classification. Our findings highlight the need for models that are capable of predicting not only the in-context polarity but also the context-autonomous polarity of aspects. Towards this goal, we introduce the CARBD-Ko dataset, a unique benchmark offering annotations for both aspect-agnostic and aspect polarities for each aspect. A distinctive feature of this dataset is the use of double-tagged aspect polarity, a detail that sets it apart from existing datasets. Furthermore, we employ a Siamese Network as an modeling approach designed to handle this double-tagged polarity. Our experimental results suggest that despite strong performance on the NSMC benchmark, the tested models face difficulty in the accurate prediction of dual-tagged polarity. This further supports our argument for the importance of models that can handle the aspect of aspects in sentiment classification. 

\section*{Ethics Statement}
The research conducted for this study was done with consideration for ethical implications and responsibilities following ACL rules. The data used for the creation of the CARBD-Ko dataset was obtained from publicly available sources, and any personally identifiable information was thoroughly removed to maintain anonymity. We have made the dataset available for research purposes only, and we trust that any subsequent use will adhere to ethical guidelines and respect the privacy and dignity of the individuals whose reviews have contributed to the dataset.

\bibliographystyle{unsrtnat}
\bibliography{references}  






\end{document}